\documentclass[lettersize,journal]{IEEEtran}

\usepackage{times}
\usepackage{epsfig}
\usepackage{graphicx}
\usepackage{amsmath}
\usepackage{amssymb}
\usepackage{stfloats}
\usepackage{multirow}
\usepackage{bbm}
\usepackage{rotating}
\usepackage{array}
\usepackage{color}
\usepackage{threeparttable}

\usepackage{epstopdf}
\usepackage{pifont}
\usepackage{subfigure}
\usepackage{bm}
\usepackage{url}
\usepackage[table]{xcolor}
\usepackage{color}
\usepackage{booktabs}
\usepackage{dsfont}
\usepackage[linesnumbered,ruled,vlined]{algorithm2e}

\begin{document}

\title{Oracle Character Recognition using Unsupervised Discriminative Consistency Network}

\author{Mei Wang, Weihong Deng, Sen Su
\thanks{Mei Wang and Weihong Deng are with the Pattern Recognition and Intelligent System Laboratory, School of Artificial Intelligence, Beijing University of Posts and Telecommunications, Beijing, 100876, China. E-mail: \{wangmei1,whdeng\}@bupt.edu.cn. (Corresponding Author: Weihong Deng)}
\thanks{Sen Su is with State Key Laboratory of Networking and Switching Technology, Beijing University of Posts and Telecommunications, Beijing, 100876, China. E-mail: susen@bupt.edu.cn.}}

\markboth{Journal of \LaTeX\ Class Files,~Vol.~14, No.~8, August~2021}%
{Shell \MakeLowercase{\textit{et al.}}: A Sample Article Using IEEEtran.cls for IEEE Journals}

\IEEEpubid{0000--0000/00\$00.00~\copyright~2021 IEEE}

\maketitle

\begin{abstract}
Ancient history relies on the study of ancient characters. However, real-world scanned oracle characters are difficult to collect and annotate, posing a major obstacle for oracle character recognition (OrCR). Besides, serious abrasion and inter-class similarity also make OrCR more challenging. In this paper, we propose a novel unsupervised domain adaptation method for OrCR, which enables to transfer knowledge from labeled handprinted oracle characters to unlabeled scanned data. We leverage pseudo-labeling to incorporate the semantic information into adaptation and constrain augmentation consistency to make the predictions of scanned samples consistent under different perturbations, leading to the model robustness to abrasion, stain and distortion. Simultaneously, an unsupervised transition loss is proposed to learn more discriminative features on the scanned domain by optimizing both between-class and within-class transition probability. Extensive experiments show that our approach achieves state-of-the-art result on Oracle-241 dataset and substantially outperforms the recently proposed structure-texture separation network by 15.1\%.
\end{abstract}


\section{Introduction}

\IEEEPARstart{A}{s} the oldest hieroglyphs in China, oracle characters \cite{flad2008divination}, which are engraved on tortoise shells and animal bones, recorded the life and history of the Shang Dynasty (around 1600-1046 B.C.) and have contributed greatly to modern civilization. To provide assistance to experts in archeology, deep convolutional neural networks (CNN) \cite{he2016deep} are recently introduced to recognize oracle characters \cite{huang2019obc306,zhang2019oracle}. However, the standard practice in deep learning, i.e., supervising with massive labeled data, becomes expensive and laborious in this field since real-world scanned oracle data are rare and data annotation requires a high level of expertise. To reduce the requirement for labeled scanned data, unsupervised domain adaptation (UDA) \cite{wang2018deep} can be a powerful approach that enables deep models to transfer knowledge from a well-labeled source domain (handprinted oracle characters) to a different but related unlabeled target domain (scanned oracle characters).

\begin{figure}[h]
\centering
\includegraphics[width=9cm]{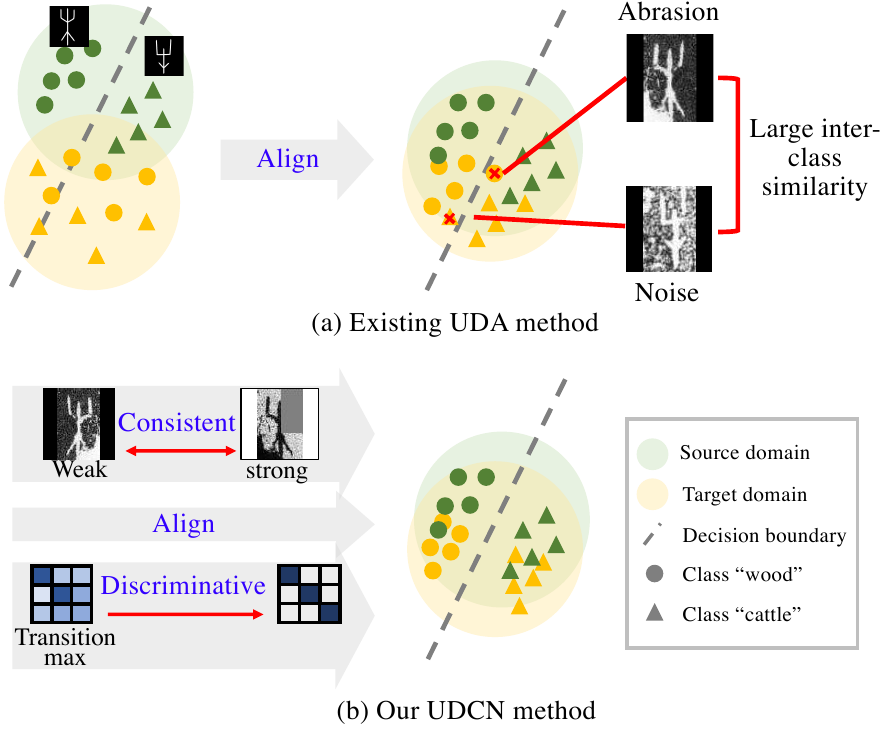}
\caption{(a) Existing UDA methods fail to be tolerant to abrasion and achieve low-density separation between classes. (b) Our UDCN learns robust and discriminative features for scanned oracle characters via augmentation consistency and unsupervised discriminative learning.}
\label{overview} 
\end{figure}
	
Commonly-used UDA methods in other classification tasks \cite{Ganin2015Unsupervised,long2018transferable} align the distributions between two domains such that the model trained on the source domain can generalize well to the target domain. However, due to the particularity of oracle characters, the two following issues should also be taken into consideration during adapting. \textbf{(1) Being robust to abrasion, stain and distortion.} Since many oracle bones have been damaged over the centuries, scanned oracle characters are fragmented and suffer from crevice, stain and noise. Besides, scanned images vary in contrast and brightness, and sometimes these characters are rotated and distorted. Domain alignment methods cannot make the model robust enough to these factors in an unsupervised manner, leading to limited improvement. \textbf{(2) Learning discriminative features.} Different writing styles lead to high degrees of intra-domain variance. Characters belonging to the same category largely vary in stroke and even topology; while some characters belonging to the different categories are similar to each other. Therefore, learning discriminative features is vital for recognizing scanned characters. \IEEEpubidadjcol

The recently proposed structure-texture separation network (STSN) \cite{9757826} utilized generative adversarial net (GAN) to transform handprinted oracle data to the scanned domain.  Although these generated data can be substitutes for unlabeled scanned data to train the network in a supervised manner such that the network can push the scanned characters of the same class together and become tolerant to abrasion and noise, the optimization of GAN suffers from instability and results in high model complexity. Besides, the features learned by cross-entropy loss are not discriminative enough in STSN. How to simply and elegantly learn robust and discriminative target features in UDA of oracle character recognition (OrCR) still mains an open problem.

In this paper, we propose a novel UDA method, called unsupervised discriminative consistency network (UDCN), which encourages augmentation consistency and inter-class discrimination on the target domain, as shown in Fig. \ref{overview}. For augmentation consistency, inspired by Fixmatch \cite{sohn2020fixmatch}, we introduce self-training (also called pseudo-labeling) \cite{saito2017asymmetric} and consistency regularization into OrCR to address the problem of lacking target labels and improve the model robustness. It generates pseudo label on a weakly augmented view of unlabeled scanned data, and then trains the network to minimize the cross entropy between the pseudo label and the model’s output on a strongly augmented view of the same instance. Pseudo-labeling enables supervised learning on real-world scanned oracle data. Instance-wise consistency enforces the model to output a consistent prediction for the different perturbed versions of the same scanned character such that the model would be less sensitive to abrasion and distortion. For inter-class discrimination, we design a new transition loss to learn discriminative target features in an unsupervised manner. We utilize the batch-wise predictions of both weakly and strongly augmented samples to compute the transition probability of an imaginary random walker starting from one target class to another. Maximizing the transition probability within a class would further enhance the batch-wise consistency between different perturbed versions. Simultaneously, minimizing the transition probability between different target classes would encourage low-density separation between classes such that characters of the same class can be pulled together and those of different classes can be pushed away from each other.

Our contributions can be summarized into three aspects.

1) UDA of oracle character recognition still remains understudied. We propose a simple yet effective UDA method, i.e., unsupervised discriminative consistency network, to learn robust and discriminative features for scanned oracle characters. It can help to unlock the cooperative potential between artificial intelligence and historians by improving the accuracy of automatic recognition systems.

2) Our proposed UDCN performs weak and strong augmentations on scanned oracle characters and enhances instance-wise consistency across different views to address the lack of labels and improve the model robustness to abrasion. Simultaneously, we optimize an unsupervised transition loss based on the transition matrix to achieve both batch-wise consistency and inter-class discrimination leading to performance improvement.

3) Extensive experimental results on Oracle-241 dataset show that our method successfully performs adaptation from handprinted oracle characters to scanned data. It substantially outperforms the recently proposed structure-texture separation network by 15.1\% and offers a new state-of-the-art (SOTA) in performance on scanned oracle characters.

\section{Related work}

\subsection{Oracle character recognition}

Oracle character recognition is one of the most fundamental aspects of oracle bone study. In the machine learning community, researches on oracle characters began in the late 1990s. Earliest works primarily extracted hand-crafted features based on graph theory and topology to recognize oracle characters. Gu et al. \cite{shaotong2016identification} encoded characters by describing the topological relations among vertexes and recognized them through topological registration. Lv et al. \cite{lv2010graphic} proposed a Fourier descriptor based on curvature histograms to encode oracle characters and perform recognition. Guo et al. \cite{guo2015building} proposed hierarchical representations for OrCR combining Gabor-related and sparse encoder-related features.

Recently, CNNs have shown considerable effectiveness in a variety of machine learning challenges, and are introduced into OrCR. Huang et al. \cite{huang2019obc306} directly trained AlexNet, VGGNet and ResNet to perform classification on scanned oracle data. OracleNet \cite{lu2020recognition} utilized a Capsule network to detect the radical-level composition and recognized oracle characters based on these radicals. Zhang et al. \cite{zhang2019oracle} trained DenseNet supervised with triplet loss and performed recognition by nearest neighbor classifier. Li et al. \cite{li2023towards} proposed a generative adversarial framework to augment oracle characters and address the problem of long-tail distribution.

However, supervised learning may be unfeasible since massive labeled oracle characters are lacking. Although UDA is an effective approach to deal with this issue, very few work has focused on UDA of OrCR. STSN \cite{9757826} is proposed to transform handprinted oracle data to the scanned domain using GAN and tune the network with these generated scanned data. In this paper, we design a simple but effective UDA method, which learns more robust and discriminative features and largely boosts the recognition performance of real-world oracle characters.

\subsection{Unsupervised domain adaptation}

Unsupervised domain adaptation \cite{wang2018deep} has received a significant amount of interest recently. A common practice for UDA is discrepancy reduction \cite{sun2016deep,ge2023unsupervised}, which aligns the distributions by moment matching. For example,  Long et al. \cite{long2018transferable} utilized the maximum mean discrepancy (MMD) to match the mean embeddings across domains. CORAL \cite{sun2016deep} aligned the second-order statistics of the source and target distributions with a linear transformation. Zellinger et al. \cite{zellinger2017central} proposed to match higher order moments by central moment discrepancy (CMD) metric.
Another effective way to achieve UDA is adversarial learning \cite{Ganin2015Unsupervised,long2018conditional}, which learns domain-invariant features in an adversarial manner. For example, DANN \cite{Ganin2015Unsupervised} trained a discriminator to distinguish the domain labels of features and simultaneously trained the feature extractor to fool the discriminator. CDAN \cite{long2018conditional} conditioned an adversarial adaptation model on discriminative information conveyed in the classifier predictions. CAADA \cite{rahman2020correlation} jointly adapted features using correlation alignment with adversarial learning.

By borrowing the idea of self-training, pseudo-label based methods \cite{saito2017asymmetric,liang2019exploring} are also widely used in UDA for most classification tasks. AsmTri \cite{saito2017asymmetric} utilized two classifiers to annotate the unlabeled target samples according to their prediction consistency, and trained another classifier by these pseudo-labeled samples. MSTN \cite{xie2018learning} applied a source classifier to generate pseudo labels for target data, and then aligned the centroids of source classes and target pseudo classes. Zou et al. \cite{zou2018unsupervised} alternatively refined pseudo labels via spatial priors and re-trained the model with these refined labels. Wang et al. \cite{wang2023improving} mined the intra-class similarity to improve the reliability of pseudo labels. In this paper, it is the first time that self-training is applied in UDA of OrCR. We combine self-training with consistency regularization to improve the model robustness, and utilize an unsupervised transition loss to further enhance inter-class discrimination.

\section{Methodology}

Assume that there are handprinted oracle characters (source domain) which consists of $n_s$ \textbf{labeled} samples $\{{x^{s}_{i}}\}^{n_s}_{i=1}$ and the corresponding labels $\{{y^{s}_{i}}\}^{n_s}_{i=1}$, as well as scanned oracle characters (target domain) which consists of $n_t$ \textbf{unlabeled} samples $\{{x^{t}_{i}}\}^{n_t}_{i=1}$. Each oracle character is located in an image. Handprinted oracle characters are written by experts, and the corresponding images are high-definition and neat. Scanned oracle data are generated by reproducing the oracle-bone surface, and suffer from abrasion, stain and distortion. Therefore, there is a discrepancy between the distributions of source and target domains. Our goal is to learn a function $f$ using $\{{x^s_i,y^{s}_{i}}\}^{n_s}_{i=1}$ and $\{{x^{t}_{i}}\}^{n_t}_{i=1}$ which can generalize well on scanned data and  predict the category of characters accurately.

\begin{figure*}
\centering
\includegraphics[width=17cm]{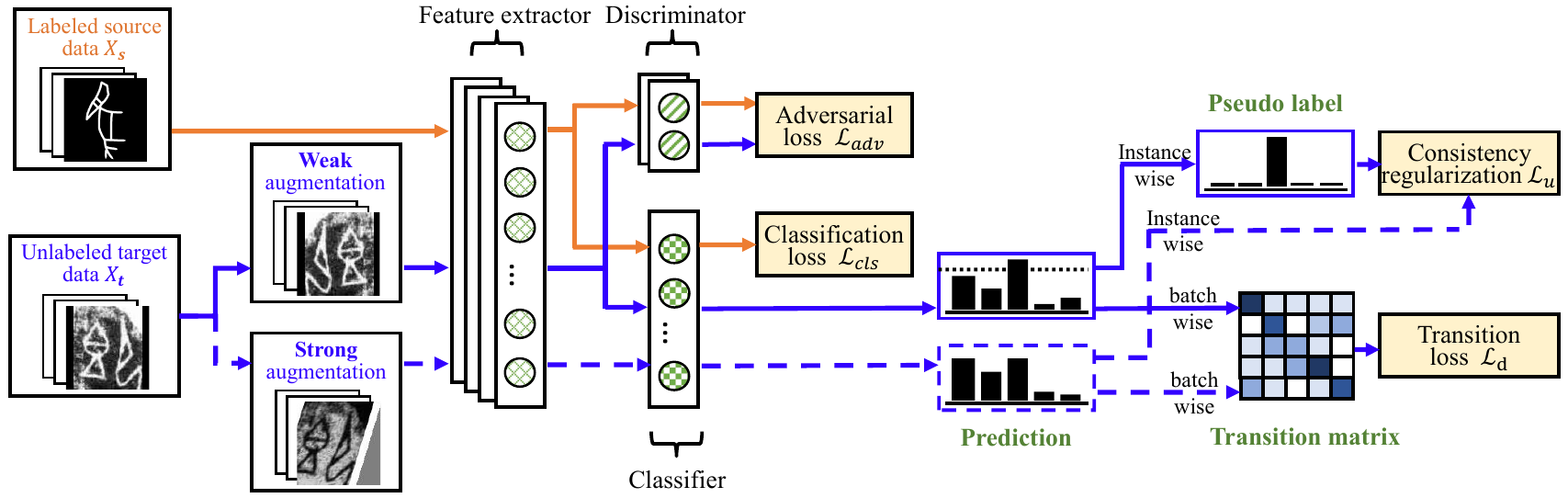}
\caption{Illustration of UDCN. We apply weak and strong augmentations to target characters and constrain the consistency of model predictions under different augmentations to improve the model robustness to abrasion and noise. Furthermore, we propose an unsupervised transition loss to minimize the between-class transition probability and maximize the within-class transition probability such that the discriminative features are learned for scanned oracle characters.}
\label{architecture} 
\end{figure*}

\subsection{Overview}

As shown in Fig. \ref{architecture}, our framework consists of a feature extractor $F$, a classifier $G$ and a discriminator $D$. We use the cross-entropy loss to calculate the loss for the source domain and employ adversarial learning to align the distribution between two domains. Pseudo-labeling and consistency regularization are performed to improve the model robustness to abrasion and noise. Specifically, two types of augmentations are applied on scanned oracle characters. We generate the pseudo label based on a weakly augmented sample, and use it as target to train on the strongly augmented version of the same sample. Furthermore, transition matrices across classes are computed based on the batch-wise predictions of different augmented views. Minimizing the between-class transition probability and maximizing the within-class transition probability can learn discriminative target features.

\subsection{Domain alignment}

Given the labeled handprinted oracle data $\{{x^s_i,y^{s}_{i}}\}^{n_s}_{i=1}$, the model can be optimized in a supervised way:
\begin{equation}
\mathcal{L}_{cls}= \mathbb{E}_{\left ( x_i^s,y_i^s \right )\sim \mathcal{D}^s} L_{CE}\left ( f\left (x_i^s\right ),y_i^s \right ) , \label{source}
\end{equation}
where $f(x_i^s)=G\left ( F\left ( x_i^s \right )  \right ) $ denotes the class prediction for $i$-th source example $x_i^s$, and $L_{CE}$ is the cross-entropy loss.

However, the model trained on handprinted oracle characters would deteriorate significantly on scanned oracle data due to the distribution discrepancy between domains. To overcome this issue, adversarial learning \cite{Tzeng2017Adversarial,Ganin2015Unsupervised} is applied, in which an additional discriminator $D$ is trained to classify whether the features extracted by $F$ are drawn from the source or target domain and simultaneously $F$ is trained to fool $D$. This min-max game is expected to reach an equilibrium where features are domain-invariant. The adversarial loss is formulated as,
\begin{equation}
\begin{split}
\underset{F}{min}\ \underset{D}{max} \ \mathcal{L}_{adv}&= \mathbb{E}_{x_i^s \sim \mathcal{D}^s} \log\left [ D\left ( F\left ( x_i^s \right )  \right )  \right ]\\
&+ \mathbb{E}_{x_i^t \sim \mathcal{D}^t} \log\left [ 1-D\left ( F\left ( x_i^t \right )  \right )  \right ]. \label{adversarial}
\end{split}
\end{equation}

\begin{figure}
\centering
\includegraphics[width=7.5cm]{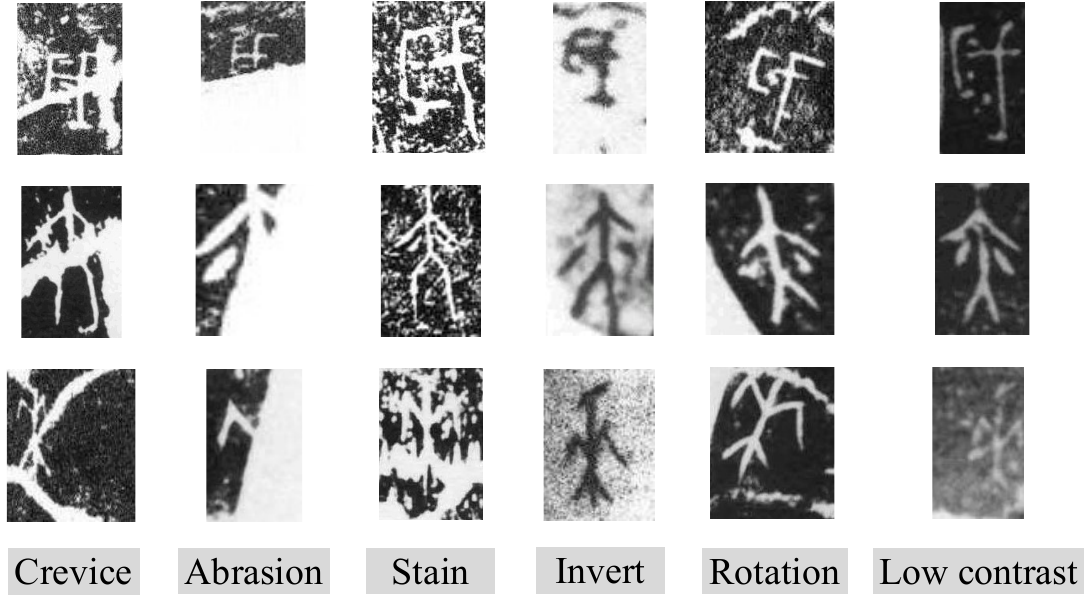}
\caption{Examples of scanned oracle characters which suffer from crevice, abrasion, stain and various kinds of distortions.}
\label{distortion} 
\end{figure}

\subsection{Pseudo-label based consistency} \label{consistency}

Domain-invariance cannot guarantee that scanned oracle characters of the same category are mapped nearby in the feature space, leading to limited improvement. Moreover, scanned characters suffer from crevice, abrasion, stain and various kinds of distortions due to long burial and careless excavation, as shown in Fig. \ref{distortion}. These factors bring great difficulty for recognition, especially when target annotations are lacking in the training process. Therefore, we generate pseudo labels as an alternative to ground-truth labels such that we are able to incorporate the semantic information into target training. Inspired by Fixmatch \cite{sohn2020fixmatch}, we further constrain the augmentation consistency on scanned data to improve the model robustness to perturbations.

\begin{figure}
\centering
\includegraphics[width=7.5cm]{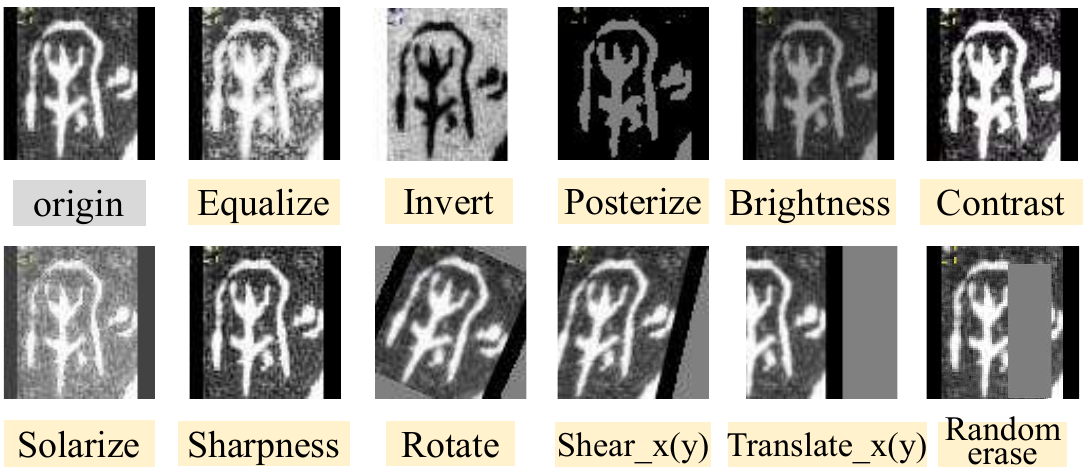}
\caption{Examples of eleven transformations in strong augmentation.}
\label{augment} 
\end{figure}

\textbf{Weak and strong augmentations.} To fully enjoy the merit of consistency training, we apply two kinds of augmentations for each target sample: ``weak'' and ``strong'', denoted by $\alpha \left ( \cdot  \right ) $ and $\mathcal{A} \left ( \cdot  \right ) $ respectively.  In our experiments, we utilize random horizontal flip as weak augmentation. Specifically, we randomly ﬂip the images with a probability of 50\%. For strong augmentation, we build a pool of operations containing eleven image transformations: Equalize, Invert, Posterize, Brightness, Contrast, Solarize, Sharpness, Rotate, Shear\_x(y), Translate\_x(y) and Random erase. We randomly apply some of them to strongly augment each sample and control the severity of the distortions by a random magnitude sampled from a pre-defined range at each training step, as shown in Table \ref{aug} and Fig. \ref{augment}.

\begin{table}
\centering
\begin{threeparttable}
\renewcommand\arraystretch{1.1}
\caption{Various transformations to strongly augment oracle characters.}
\label{aug}
    \setlength{\tabcolsep}{4mm}{
	\begin{tabular}{c|ccc}
		\toprule
         Operation & Equalize & Invert & Posterize \\
         Range & 0 or 1	& 0 or 1 & [4,8] \\ \hline
         Operation & Brightness & Contrast & Solarize \\
         Range & [0.05,0.95] & [0.05,0.95] & [0,1] \\  \hline
         Operation & Sharpness & Rotate & Shear\_x(y) \\
         Range & [0.05,0.95] & [-30,30] & [-0.3,0.3] \\ \hline
         Operation & Translate\_x(y) & \multicolumn{2}{c}{Random erase$^*$} \\
         Range & [-0.3,0.3]& \multicolumn{2}{c}{$sl$=0.02, $sh$=0.4, $r1$=0.3} \\
         \bottomrule
    \end{tabular}}
    \begin{tablenotes}
    \scriptsize
        \item[*] The position and area of the erasing rectangle region are randomly sampled. $sl$ and $sh$ define the minimum and maximum area of the erased rectangle, and $r1$ denotes the aspect ratio of the erased area.
     \end{tablenotes}
\end{threeparttable}
\end{table}

The reason for using random horizontal flip as weak augmentation and using others as strong augmentation is as follows. Oracle characters are insensitive to flipping. There are many characters that mirror each other in both handprinted and scanned domains. Therefore, applying horizontal flip makes little change to oracle characters. On the contrary, other transformations, e.g., Brightness, Equalize and Random erase, aim to artificially simulate the abrasions in the real world which makes the augmented characters look more like rubbings. Additionally, the magnitude of these transformations is sampled randomly and sometimes can be significant enough to make the augmentation strong.

\textbf{Pseudo labeling.} For a weakly augmented version of scanned oracle character $x_i^t$, we pick up the class with the maximum predicted probability as its pseudo label $\hat{y}_i^t$,
\begin{equation}
\hat{y}_i^t=\begin{cases}
 & \arg\max p_i^\alpha, \ \ \ \max p_i^\alpha> \tau, \\
 &-1, \ \ \ \ \ \ \ \ \ \ \ \text{ otherwise, }  \label{pseudo}
\end{cases}
\end{equation}
where $p_i^\alpha=f\left ( \alpha \left ( x_i^t \right ) \right )$ is the model prediction for a weakly augmented version of $x_i^t$. To mitigate the negative effect caused by falsely-labeled samples, we only assign pseudo labels for the samples whose prediction confidences are above the threshold $\tau$. Therefore, the label ` -1' means that low-confident samples are abandoned and would not be involved in the training of augmentation consistency.

\textbf{Augmentation consistency.} After generating the pseudo labels, we utilize them as targets to train on the strongly augmented views of these unlabeled images,
\begin{equation}
\mathcal{L}_{u}= \mathbb{E}_{ x_i^t \sim \mathcal{D}^t} L_{CE}\left ( f\left (\mathcal{A} \left (x_i^t  \right )\right ),\hat{y}_i^t \right ).
\end{equation}
Pseudo-labeling can address the lack of target labels and enable supervise learning on scanned oracle characters, which encourages the separability of features. Simultaneously, augmentation consistency makes the predictions of scanned samples consistent under different perturbations, leading to the model robustness to various kinds of distortions.

\begin{figure}
\centering
\subfigure[` horse' ]{
\label{horse} 
\includegraphics[height=2.6cm]{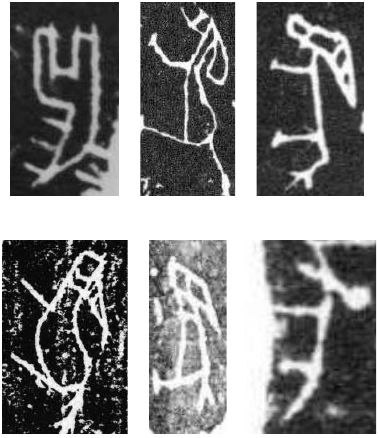}}
\hspace{0.3cm}
\subfigure[` wood']{
\label{wood} 
\includegraphics[height=2.6cm]{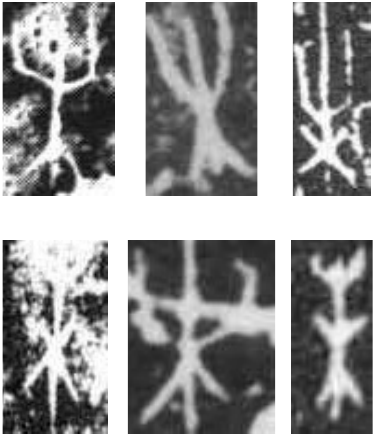}}
\hspace{0.3cm}
\subfigure[` cattle']{
\label{cattle} 
\includegraphics[height=2.6cm]{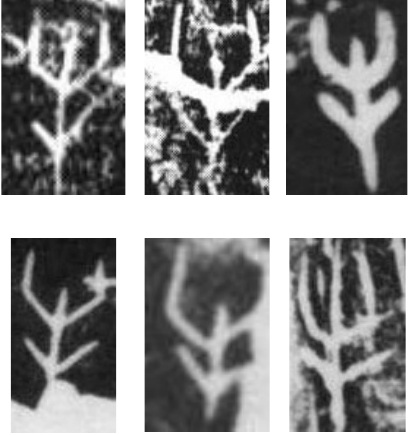}}
\caption{Examples from the ` horse', ` wood' and ` cattle' categories. Different writing styles lead to a high degree of intra-class variance and inter-class similarity.}
\label{example character} 
\end{figure}

\subsection{Unsupervised discriminative learning} \label{training loss}

For OrCR task, the deeply learned features need to be not only separable but also discriminative. Oracle characters evolved over time which results in various writing styles. As shown in Fig. \ref{horse}, characters belonging to the same category largely vary in stroke and even topology, leading to serious intra-class variance. Some characters belonging to the different categories are similar to each other, which may confuse the recognition model. For example, the inter-class similarity between the characters of `wood' and `cattle' is quite large shown in Fig. \ref{wood} and \ref{cattle}. Therefore, both the compact intra-class variations and separable inter-class differences are of vital importance for improving the performance on scanned oracle characters. To address this issue, we propose an unsupervised transition loss.

\begin{figure}
\centering
\includegraphics[width=8cm]{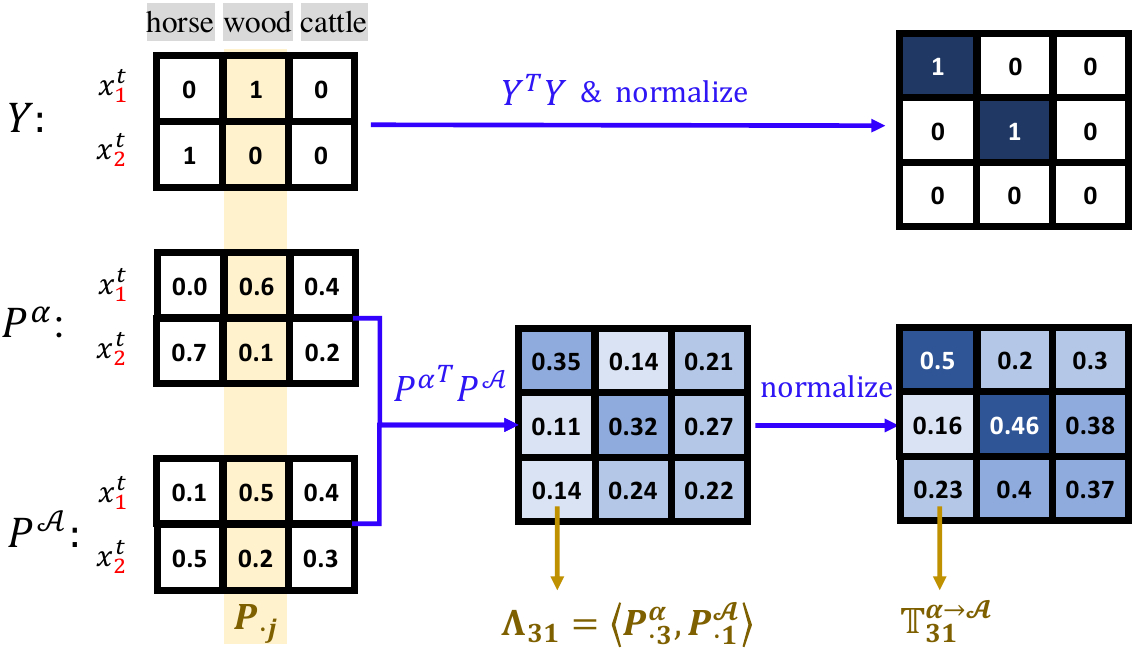}
\caption{Transition probability across classes within target domain.}
\label{transition} 
\end{figure}

\textbf{Transition probability.} As shown in Fig. \ref{transition}, assume that $P^\alpha=f\left ( \alpha\left ( X^t \right ) \right ) \in \mathbb{R}^{B\times C}$ denotes the model predictions on a batch of target data with weak augmentation, where $B$ is the batch size and $C$ is the category number. Accordingly, the $i$-th column of $P^\alpha$, denoted by $P^\alpha_{\cdot i}\in \mathbb{R}^{B\times 1}$, reveals the probabilities that $B$ samples in each batch come from the $i$-th class. Similarly, $P^{\mathcal{A}}_{\cdot j}\in \mathbb{R}^{B\times 1}$ denotes the probabilities that the strongly augmented views of $B$ samples belong to the $j$-th class. We compute the dot product between $P^{\alpha }_{\cdot i}$ and $P^{\mathcal{A}}_{\cdot j}$ to coarsely estimate the class-wise correlation (confusion) between $i$-th and $j$-th classes under different views, which is formulated as,
\begin{equation}
\Lambda _{ij}=\left \langle P^{\alpha }_{\cdot i}\cdot P^{\mathcal{A}}_{\cdot j} \right \rangle,
\end{equation}
where $\left \langle \cdot  \right \rangle$ is the operation of dot product. It measures the possibility that the trained model simultaneously classifies the $B$ samples into the $i$-th class under weakly augmented view and into the $j$-th class under strongly augmented view. It is noteworthy that $\Lambda _{ij}\neq \Lambda _{ji}$.

If one considers the transitions of an imaginary random walker between two classes, the intuition is that transitions are more probable when they are more correlated and similar. Therefore, we normalize $\Lambda _{ij}$ to approximate the transition probability from $i$-th class to $j$-th class,

\begin{equation}
\mathbb{T}^{\alpha \rightarrow \mathcal{A}}_{ij}=\mathbb{P}\left ( \mathcal{A}_j|\alpha_i \right ) =\frac{\Lambda _{ij}}{\sum_{j=1}^{C}\Lambda _{ij}},
\end{equation}
\begin{equation}
\mathbb{T}^{\mathcal{A} \rightarrow \alpha}_{ij}=\mathbb{P}\left ( \alpha_j|\mathcal{A}_i \right )=\frac{\Lambda _{ji}}{\sum_{j=1}^{C}\Lambda _{ji}},
\end{equation}
where $\mathbb{T}^{\alpha \rightarrow \mathcal{A}}_{ij}$ is the transition probability of an imaginary random walker starting from $i$-th weakly augmented class to $j$-th strongly augmented class, and $\mathbb{T}^{\mathcal{A} \rightarrow \alpha}_{ij}$ is the transition probability from $i$-th strongly augmented class to $j$-th weakly augmented class.

\textbf{Discriminative learning.} With the cluster assumption that samples in the same class should form a high-density cluster, we hope to encourage the class-wise correlation $\Lambda _{ij}$ equals to 0 when $i\neq j$ as well as $\Lambda _{ij}\geq 0$ when $i=j$. Namely, on the one hand, the classifier should not ambiguously classify samples into two different classes at the same time no matter what perturbation is performed. On the other hand, weakly and strongly augmented views of the same samples should be confidently classified into the same class. Based on this basic assumption, we propose to learn discriminative features by optimizing the transition probabilities $\mathbb{T}^{\alpha \rightarrow \mathcal{A}}_{ij}$ and $\mathbb{T}^{\mathcal{A} \rightarrow \alpha}_{ij}$,
\begin{equation}
\begin{split}
\mathcal{L}_{d}&=\frac{1}{2}\left ( \sum_{i=1}^{C} \sum_{j\neq i}^{C}\mathbb{T}^{\alpha \rightarrow \mathcal{A}}_{ij}- \sum_{i=1}^{C} \mathbb{T}^{\alpha \rightarrow \mathcal{A}}_{ii}\right )\\
&+\frac{1}{2}\left ( \sum_{i=1}^{C} \sum_{j\neq i}^{C}\mathbb{T}^{\mathcal{A} \rightarrow \alpha}_{ij} -\sum_{i=1}^{C}\mathbb{T}^{\mathcal{A} \rightarrow \alpha}_{ii}\right ). \label{discriminative}
\end{split}
\end{equation}
Minimizing the \emph{between-class} transition probabilities $\sum_{i=1}^{C} \sum_{j\neq i}^{C}\mathbb{T}^{*}_{ij}$ can decrease the correlation between different classes such that discriminative features with large inter-class differences and small intra-class variations are learned. Maximizing the \emph{within-class} transition probabilities $\sum_{i=1}^{C} \mathbb{T}^{*}_{ii}$ would further enhance the batch-wise consistency between samples with different perturbations and thus improve the model robustness. Constraining the consistency along the batch dimension means that we view unlabeled data as ensemble rather than individual sample. It is advantageous to eliminate anomalous sample disturbance when the ground-truth labels are lacking. Moreover, as we described in Section \ref{consistency}, only samples with high confidence can be assigned pseudo labels and utilized to optimize the model in pseudo-label based consistency $\mathcal{L}_{u}$, which would unfortunately miss some useful information contained in the remaining unlabeled data. However, $\mathcal{L}_{d}$ is an unsupervised loss by which we can take full advantage of the whole data and compensate for the weakness of $\mathcal{L}_{u}$.

\subsection{Overall objective}

Combining the source classification loss, adversarial loss, consistency regularization and transition loss, our overall objective is formulated as:
\begin{equation}
\begin{split}
&\underset{F}{min}\ \mathcal{L}_{cls}+\mathcal{L}_{adv}+\mathcal{L}_{u}+\lambda \mathcal{L}_{d}, \\
&\underset{D}{max}\ \mathcal{L}_{adv},
\end{split}
\end{equation}
where $\lambda$ is the trade-off parameter to balance losses. Note that adversarial loss $\mathcal{L}_{adv}$ is computed on the weakly augmented views of target samples.

\section{Experiments}

We evaluate the proposed UDCN on the benchmarks of OrCR as well as digit classification to prove its effectiveness on transferring knowledge across domains and compare UDCN with the SOTA domain adaptation methods.

\subsection{Datasets}

\begin{figure*}
\centering
\subfigure[Handprinted oracle characters]{
\label{hand_ex} 
\includegraphics[height=4.3cm]{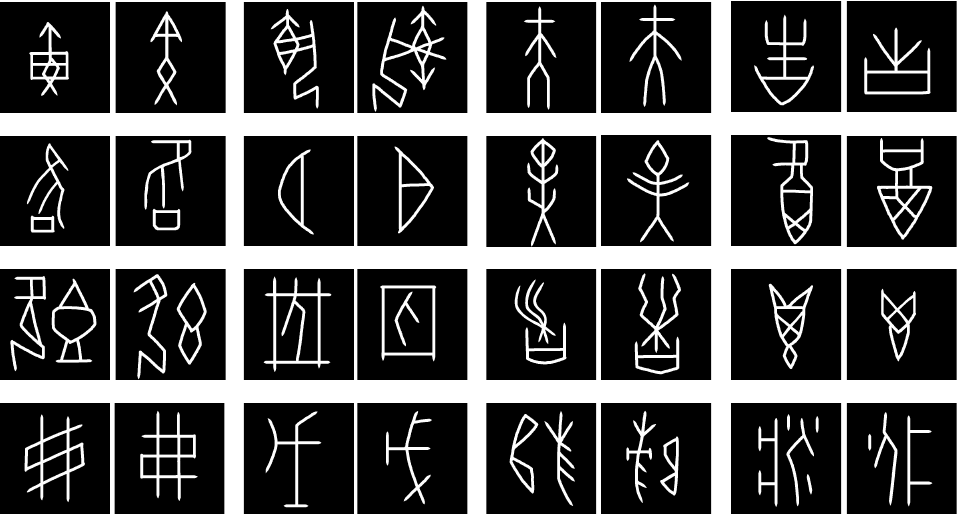}}
\hspace{0.5cm}
\subfigure[Scanned oracle characters]{
\label{scan_ex} 
\includegraphics[height=4.3cm]{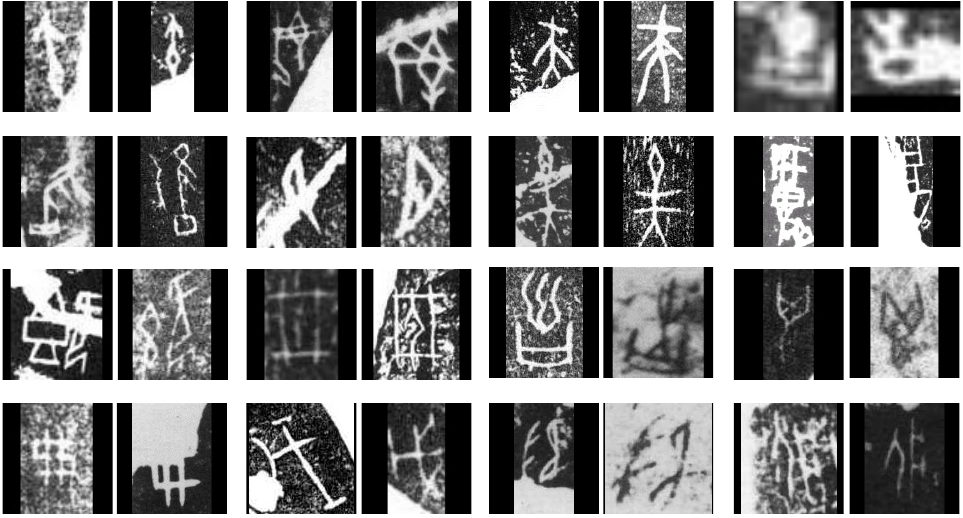}}
\caption{Examples of (a) handprinted and (b) scanned characters in Oracle-241 dataset.}
\label{Oracle241} 
\end{figure*}

\textbf{Oracle-241} \cite{9757826} is an oracle character dataset\footnote{Oracle-241 dataset is available on https://github.com/wm-bupt/STSN.} for domain adaptation which contains 80K handprinted and scanned oracle characters belonging to 241 categories in total, as shown in Fig. \ref{Oracle241}. Each character is located in an image. We follow \cite{9757826} to split the dataset into training and testing sets. We utilize labeled 10,861 handprinted data and unlabeled 50,168 scanned data for training, and use 3,730 handprinted data and 13,806 scanned data for testing.

\textbf{MNIST-USPS-SVHN \cite{lecun1998gradient,netzer2011reading,denker1989neural}} are commonly-used UDA datasets containing 10 classes of digits. MNIST and USPS are both handwritten digit datasets. MNIST consists of 70K greyscale images and USPS dataset includes 9.3K greyscale images. SVHN is a real-world digits dataset obtained from house numbers in Google street view images, which consists of 100K images. Following \cite{hoffman2018cycada}, we evaluate the methods on three transfer tasks: MNIST$\rightarrow$USPS (M$\rightarrow$U), USPS$\rightarrow$MNIST (U$\rightarrow$M) and SVHN$\rightarrow$MNIST (S$\rightarrow$M).

\subsection{Implementation detail}

\textbf{Network architecture.} All the experiments are implemented with PyTorch. For OrCR, we use ResNet-18 \cite{he2016deep} pretrained on ImageNet as our backbone network and replace the last FC layer with the task-specific FC layer. For digit classification, we use a variant of the LeNet architecture as backbone following \cite{hoffman2018cycada,long2018conditional}. The same discriminator architecture is utilized as DANN \cite{Ganin2015Unsupervised} and we apply gradient reversal layer (GRL) \cite{Ganin2015Unsupervised} to achieve the min-max optimization in adversarial learning.

\textbf{Experimental setup.} We set the trade-off parameter $\lambda$ to be 1 and set the threshold $\tau$ to be 0.95. We use mini-batch stochastic gradient descent (SGD) with momentum of 0.9 to train the network. For OrCR task, we train the model for 50,000 iterations with the batch size of 32. We follow the same learning rate schedule as described in \cite{9757826}. The learning rate is annealed by $\eta=\eta_0\left ( \frac{1-T}{T_{max}} \right )^{0.9}$, where $\eta_0=0.001$ is the initial learning rate, $T$ and $T_{max}$ are the current and total iteration. For digit classification, we train the model for 80 epochs with the batch size of 64. The learning rate is set to be 0.01 and divided by 10 at 40 epochs in U$\rightarrow$M and M$\rightarrow$U tasks. In S$\rightarrow$M task, we set the learning rate to be 0.003 and divide it by 10 at 40 epochs.

\begin{table}
\renewcommand\arraystretch{1.1}
\caption{Source and target accuracy (mean$\pm$std\%) on Oracle-241 dataset. The best accuracy is
indicated in \textbf{\textcolor{red}{bold red}} and the second best accuracy is indicated in \textcolor{blue}{\underline{undelined blue}}.}
 \label{oracle-241}
	\begin{center}
   \setlength{\tabcolsep}{3.5mm}{
	\begin{tabular}{c|l|ccc}
		\toprule
           \multicolumn{2}{c|}{Methods} & Source: & Target: \\
         \multicolumn{2}{c|}{ } & Handprint & Scan  \\ \hline \hline
         Source-  & ResNet (ICDAR19) \cite{huang2019obc306} & \textcolor{blue}{\underline{94.9$\pm $0.1}}  &  2.1$\pm $0.6   \\
         only &NN-DML (ICDAR19) \cite{zhang2019oracle} & 94.5$\pm $0.4  &  8.4$\pm $1.0   \\\hline \hline
         \multirow{14}{*}{UDA}& CORAL (ECCV16) \cite{sun2016deep} & 89.5$\pm $0.6 &  18.4$\pm $1.3 \\
         & DDC (ArXiv14) \cite{tzeng2014deep} & 90.8$\pm $1.5  &   25.6$\pm $1.9   \\
         & DAN (TPAMI18) \cite{long2018transferable} & 90.2$\pm $1.5  &   28.9$\pm $1.6   \\
         & ASSDA (TIP21) \cite{zhang2021robust} &85.8$\pm $0.1 &32.6$\pm $0.2 \\
         & DANN (ICML15) \cite{Ganin2015Unsupervised} & 87.1$\pm $1.7 &  32.7$\pm $1.5     \\
         & GVB (CVPR20) \cite{cui2020gradually} & 92.8$\pm $0.4  & 36.8$\pm $1.1  \\
         & CDAN (NeurIPS18) \cite{long2018conditional} & 85.3$\pm $3.3 & 37.9$\pm $2.0  \\
         & MSTN (ICML18) \cite{xie2018learning} & 91.0$\pm $1.5 & 38.3$\pm $1.2 \\
         & TransPar (TIP22) \cite{9807644} & 93.1$\pm $0.4 & 39.8$\pm $1.1 \\
         & FixBi (CVPR21) \cite{na2021fixbi} & 90.1$\pm $1.6 & 40.2$\pm $0.1 \\
         & PRONOUN (TIP21) \cite{hu2021adversarial} & 92.4$\pm $0.3 & 40.3$\pm $1.8  \\
         & BSP (ICML19) \cite{chen2019transferability} &87.7$\pm $0.7 & 43.7$\pm $0.4  \\
         & STSN (TIP22) \cite{9757826}  & \textbf{\textcolor{red}{95.0$\pm $0.2}} &  \textcolor{blue}{\underline{47.1$\pm$0.8}}   \\ \cline{2-4}
         & \cellcolor{blue!5}\textbf{UDCN (ours)}  & \cellcolor{blue!5}93.6$\pm $0.2 & \cellcolor{blue!5}\textbf{\textcolor{red}{62.2$\pm$0.5}}   \\
         \bottomrule
         \end{tabular}}
    \end{center}
\end{table}

\textbf{Evaluation protocol.} We follow the standard protocols for unsupervised domain adaptation as \cite{Ganin2015Unsupervised}.  For each method, we run three random experiments and report the average classification accuracy and the standard deviation. In digit classification task, we adopt the result on target domain as evaluation metric. In OrCR task, we measure the results on both source and target domains.

\subsection{Comparison with state-of-the-arts}

\textbf{Results on Oracle-241.} Table \ref{oracle-241} reports the results on Oracle-241 dataset to evaluate the effectiveness of our method on transferring recognition knowledge from handprinted oracle characters to scanned data. In addition to UDA methods, we also compare our UDCN with the methods designed for OrCR, e.g., NN-DML  \cite{zhang2019oracle}. We call them `source-only' models, since they only train models on handprinted data without adaptation. For fair comparison, we re-implement all of the comparison methods on Oracle-241 using the same backbone and training protocol including batch size.

From the results, we have the following observations. (1) Without adaptation, `source-only' models cannot perform well on scanned data, which proves the existence of the distribution discrepancy between handprinted and scanned data. (2) The target performance can be improved but is still less than 45\% when existing UDA methods designed for object classification are introduced into OrCR. For example, with the help of self-training, MSTN \cite{long2018conditional} beats NN-DML in terms of target accuracy but only obtains 38.3\%. TransPar \cite{9807644} performed separate update rules on transferable and untransferable parameters to reduce the side effect of domain-specific information. However, this implicit operation cannot make the model robust and discriminative enough. (3) STSN \cite{9757826} published in TIP'22 is the first work focusing on UDA of OrCR. Benefitting from joint disentanglement, transformation and adaptation, it surpasses the best UDA competitor, i.e., BSP \cite{chen2019transferability}, by 3.4\%. (4) Our proposed UDCN achieves the best average accuracy on scanned data compared with all the baselines, and even significantly outperforms STSN by 15.1\%. Due to the large distribution discrepancy, it is difficult for UDA methods to simultaneously achieve high performances on both domains. Although the source performance of UDCN is slightly decreased compared with `source-only' methods since UDCN pays more attention to the target domain, it remains to be 93.6\% and achieves the second-best result among the comparison UDA methods.

\begin{table*}
\renewcommand\arraystretch{1.1}
\caption{Target accuracy (mean$\pm$ std\%) on three transfer tasks of digit datasets. }
\label{MNIST-USPS-SVHN}
	\begin{center}
    \setlength{\tabcolsep}{3.4mm}{
	\begin{tabular}{l|ccc|c}
		\toprule
         Methods &   U$\rightarrow $M  & M$\rightarrow $U & S$\rightarrow $M & \cellcolor{gray!7} Avg \\ \hline \hline
         Source-only (CVPR19) \cite{he2016deep} & 69.6$\pm$3.8 & 82.2$\pm$0.8 & 67.1$\pm$0.6 & \cellcolor{gray!7} 73.0 \\
         DANN (ICML15) \cite{Ganin2015Unsupervised} & - & 77.1$\pm$1.8 & 73.6& \cellcolor{gray!7} - \\
         DRCN (ECCV16) \cite{ghifary2016deep} & 73.7$\pm$0.1 & 91.8$\pm$0.1 & 82.0$\pm$0.2 & \cellcolor{gray!7} 82.5 \\
         ADDA (CVPR17) \cite{Tzeng2017Adversarial} & 90.1$\pm$0.8 & 89.4$\pm$0.2 & 76.0$\pm$1.8 & \cellcolor{gray!7} 85.2 \\
         DAA (Neuro19) \cite{jia2019domain} &  92.8$\pm$1.1 & 90.3$\pm$0.2 & 78.3$\pm$0.5 & \cellcolor{gray!7} 87.1 \\
         LEL (NeurIPS17) \cite{luo2017label} & - & - & 81.0$\pm$0.3& \cellcolor{gray!7} - \\
         DSN (NeurIPS16) \cite{bousmalis2016domain} & - & - & 82.7& \cellcolor{gray!7} - \\
         DTN (ICLR17) \cite{taigman2016unsupervised} & - & - & 84.4& \cellcolor{gray!7} - \\
         AsmTri (ICML17) \cite{saito2017asymmetric} &   - & - & 86.0& \cellcolor{gray!7} - \\
         CoGAN (NeurIPS16) \cite{liu2016coupled}  & - & - & 91.2$\pm$0.8& \cellcolor{gray!7} - \\
         MSTN (ICML18) \cite{xie2018learning} & - & 92.9$\pm$1.1 & 91.7$\pm$1.5& \cellcolor{gray!7} - \\
         PixelDA (CVPR17) \cite{bousmalis2017unsupervised} & & \textcolor{blue}{\underline{95.9}} & & \cellcolor{gray!7} -\\
         TPN  (CVPR19) \cite{pan2019transferrable} & 94.1 &  92.1 & 93.0 & \cellcolor{gray!7}  93.1 \\
         UNIT (NeurIPS17) \cite{liu2017unsupervised} & 93.6 & \textcolor{blue}{\underline{95.9}} & 90.5 & \cellcolor{gray!7} 93.4\\
         CyCADA (ICML18) \cite{hoffman2018cycada} & 96.5$\pm$0.1 & 95.6$\pm$0.2 & 90.4$\pm$0.4 & \cellcolor{gray!7} 94.2 \\
         CDAN (NeurIPS18) \cite{long2018conditional} & \textcolor{blue}{\underline{98.0}} & 95.6 &  89.2 &  \cellcolor{gray!7} 94.3 \\
         STSN (TIP22) \cite{9757826} & 96.7$\pm$0.1 &  94.4$\pm$0.3 &  92.2$\pm$0.1 & \cellcolor{gray!7} \textcolor{blue}{\underline{94.4}} \\
         PFAN (CVPR19) \cite{chen2019progressive} & - & 95.0$\pm$1.3 & \textcolor{blue}{\underline{93.9$\pm$0.8}} & \cellcolor{gray!7} - \\
         \hline \rowcolor{blue!5}
         \textbf{UDCN (ours)} &  \textcolor{red}{\textbf{98.3$\pm$0.1}} &  \textcolor{red}{\textbf{96.2$\pm$0.1}} &  \textcolor{red}{\textbf{95.8$\pm$1.0}} &\cellcolor{gray!7} \textcolor{red}{\textbf{96.8}}  \\
         \bottomrule
	\end{tabular}}
    \end{center}
\end{table*}

\begin{table}
\renewcommand\arraystretch{1.1}
\small
\caption{Ablation investigations of our model on Oracle-241 dataset. ACC means the accuracy on scanned data, and $\Delta$ denotes the increase relative to the previous row.}
\label{ablation}
	\begin{center}
    \setlength{\tabcolsep}{4mm}{
	\begin{tabular}{cccc|cc}
		\toprule
         $\mathcal{L}_{cls}$  & $\mathcal{L}_{adv}$  & $\mathcal{L}_{u}$ & $\mathcal{L}_{d}$  & ACC & $\Delta$(\%) \\ \hline \hline
         \ding{51}& \textcolor{red}{\ding{55}} & \textcolor{red}{\ding{55}} & \textcolor{red}{\ding{55}}   &2.1 & -  \\
         \ding{51}& \ding{51} & \textcolor{red}{\ding{55}} & \textcolor{red}{\ding{55}}  & 32.7 & $\uparrow$30.6  \\
         \ding{51}&  \ding{51} & \ding{51} & \textcolor{red}{\ding{55}}  & 52.6 & $\uparrow$19.9   \\
         \ding{51}& \ding{51}& \ding{51} & \ding{51}  & 62.2 & $\uparrow$ 9.6  \\
         \bottomrule
         \end{tabular}}
    \end{center}
\end{table}

\textbf{Results on MNIST-USPS-SVHN.} Table \ref{MNIST-USPS-SVHN} reports the results on digit datasets to prove that UDCN has the potential to generalize to a variety of scenarios. For each transfer task, we report the target accuracies provided by the original papers. (1) From the results, we can see that UDCN performs better on digit datasets compared with Oracle-241 dataset, proving the difficulty of OrCR. The domain gap in Oracle-241 is much larger than that in MNIST-USPS-SVHN which makes it difficult for UDA methods to adapt. Besides, large intra-class variance and high inter-class similarity in Oracle-241 confuse the recognition model leading to poorer performance even for handprinted oracle characters. (2) Our method is able to outperform previous UDA methods on digit datasets.  Specifically, it achieves 98.3\%, 96.2\% and 95.8\% on U$\rightarrow$M, M$\rightarrow$U and S$\rightarrow$M, respectively, and boosts the average accuracy of CDAN \cite{long2018conditional} by 2.5\% and that of STSN \cite{9757826} by 2.4\%. The results further verify the advances of our method. Compared with class-aware alignment method, e.g., CDAN \cite{long2018conditional}, the utilization of class information is more explicit and effective in our UDCN. Compared with other self-training methods, e.g., MSTN \cite{xie2018learning}, AsmTri \cite{saito2017asymmetric}, and PFAN \cite{chen2019progressive}, our method advances from two aspects. The first is that UDCN constrains the consistency of predictions under different perturbations which makes the network more robust. The other is that our method utilizes an unsupervised transition loss to encourage the features to be not only separable but also discriminative.

\subsection{Ablation study}

\textbf{Effectiveness of each component.} We report the ablation analysis of different losses in our UDCN via extensive variant experiments on Oracle-241 dataset. The accuracies on scanned data are shown in Table \ref{ablation}. We denote the method only using $\mathcal{L}_{cls}$ as \emph{BASE}. From the results, we can find that each component makes an important contribution to the target performance. (1) Pseudo-label based consistency $\mathcal{L}_{u}$. We can observe that \emph{BASE}+$\mathcal{L}_{adv}$+$\mathcal{L}_{u}$ significantly outperforms \emph{BASE}+$\mathcal{L}_{adv}$ by 19.9\%, confirming the effectiveness of self-training and augmentation consistency. Self-training enables the network to match the distributions for each category and augmentation consistency makes the network tolerant to abrasion and noise, which guarantees better adaptation.
(2) Unsupervised transition loss $\mathcal{L}_{d}$. When $\mathcal{L}_{d}$ is abandoned, the adaptation performance degrades 9.6\% in terms of target accuracy. $\mathcal{L}_{d}$ constructs the transition matrices across classes and decreases the class-wise correlation leading to the discriminative features. Moreover, it further enhances the batch-wise consistency between different augmented views and utilizes the missing information contained in low-confident samples which compensates for the weakness of $\mathcal{L}_{u}$.

\begin{table}
\renewcommand\arraystretch{1.1}
\small
\caption{Effectiveness of weak and strong augmentations on Oracle-241 dataset. ACC means the accuracy on scanned data, and $\Delta$ denotes the increase relative to the previous row.}
\label{ablation_WS}
	\begin{center}
    \setlength{\tabcolsep}{1.8mm}{
	\begin{tabular}{l|cc|cc|cc}
		\toprule
         \multirow{2}{*}{Method} & \multicolumn{2}{c|}{$\mathcal{L}_{u}$}  &  \multicolumn{2}{c|}{$\mathcal{L}_{d}$} & \multirow{2}{*}{ACC} & \multirow{2}{*}{$\Delta$(\%)} \\
          &  Weak  & Strong & Weak  & Strong & & \\ \hline \hline
         $\mathcal{L}_{u}$ \emph{w/ W} &\ding{51} & \textcolor{red}{\ding{55}} & \textcolor{red}{\ding{55}} & \textcolor{red}{\ding{55}}  & 45.3 & -  \\
          $\mathcal{L}_{u}$ \emph{w/ WS} &\ding{51}  & \ding{51} & \textcolor{red}{\ding{55}}  & \textcolor{red}{\ding{55}} & 52.6 & $\uparrow$ 7.3  \\ \hline
          $\mathcal{L}_{d}$ \emph{w/ W} &\ding{51}  & \ding{51}  &  \ding{51} & \textcolor{red}{\ding{55}}  & 59.5 & -   \\
          $\mathcal{L}_{d}$ \emph{w/ WS} &\ding{51}  & \ding{51}  & \ding{51} & \ding{51}  & 62.2 & $\uparrow$ 2.7  \\
         \bottomrule
         \end{tabular}}
    \end{center}
\end{table}

\textbf{Effectiveness of weak and strong augmentations.} As illustrated in Table \ref{ablation_WS}, we compare the proposed method trained with and without strong augmentation. (1) Pseudo-label based consistency $\mathcal{L}_{u}$. $\mathcal{L}_{u}$ \emph{w/ W} performs traditional self-training on weakly augmented samples; while $\mathcal{L}_{u}$ \emph{w/ WS} combines self-training with consistency regularization based on both weakly and strongly augmented views. From the results, $\mathcal{L}_{u}$ \emph{w/ WS} improves the target accuracy by 7.3\% over $\mathcal{L}_{u}$ \emph{w/ W}, demonstrating the effectiveness of the consistency regularization across views. (2) Unsupervised transition loss $\mathcal{L}_{d}$. $\mathcal{L}_{d}$ \emph{w/ WS} constructs the transition matrices based on both weakly and strongly augmented views, while $\mathcal{L}_{d}$ \emph{w/ W} is only based on weakly augmented samples. It can be seen that $\mathcal{L}_{d}$ \emph{w/ WS} outperforms $\mathcal{L}_{d}$ \emph{w/ W}. We believe the reason is as follows. $\mathcal{L}_{d}$ \emph{w/ W} plays the similar role with entropy minimization \cite{grandvalet2004semi} which encourages the low-density separation between classes. However, it ignores the importance of maintaining discriminability and reliability under noise and abrasion. Our $\mathcal{L}_{d}$ \emph{w/ WS} can reduce class confusion no matter what perturbation is performed.

\subsection{Parameter sensitivity}

\begin{figure*}
\centering
\subfigure[Oracle-241: handprint$\rightarrow $scan]{
\label{hand_ex} 
\includegraphics[height=3.1cm]{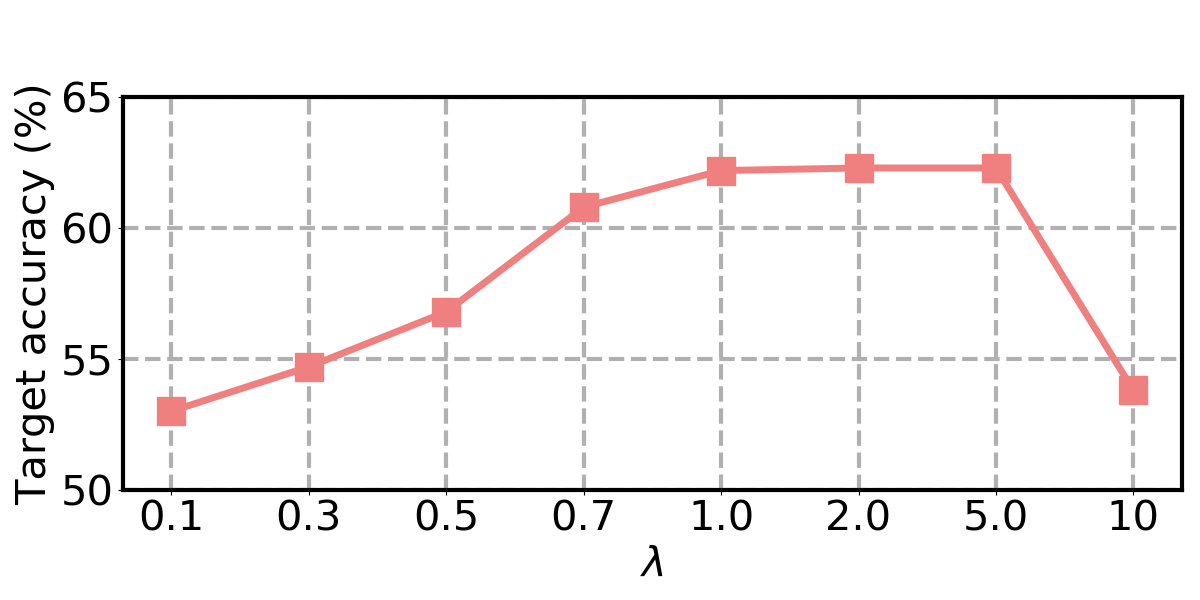}}
\hspace{0cm}
\subfigure[MNIST-USPS-SVHN: U$\rightarrow $M]{
\label{scan_ex} 
\includegraphics[height=3.1cm]{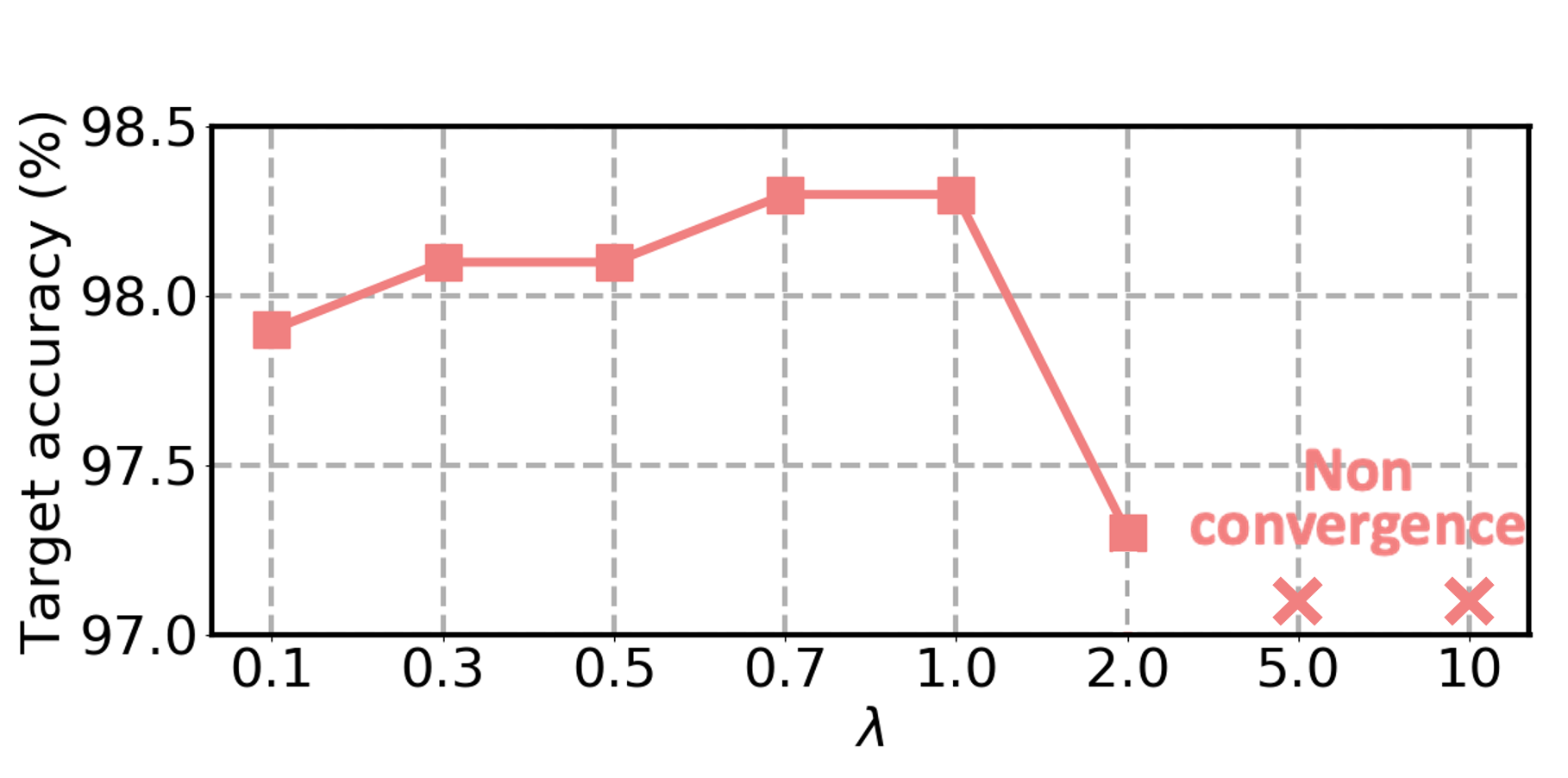}}
\caption{Parameter sensitivity investigation of $\lambda$ in terms of target accuracy on (a) Oracle-241 and (b) MNIST-USPS-SVHN datasets.}
\label{sensitivity} 
\end{figure*}

\begin{figure*}
\centering
\subfigure[Mask rate]{
\label{hand_ex} 
\includegraphics[height=2.8cm]{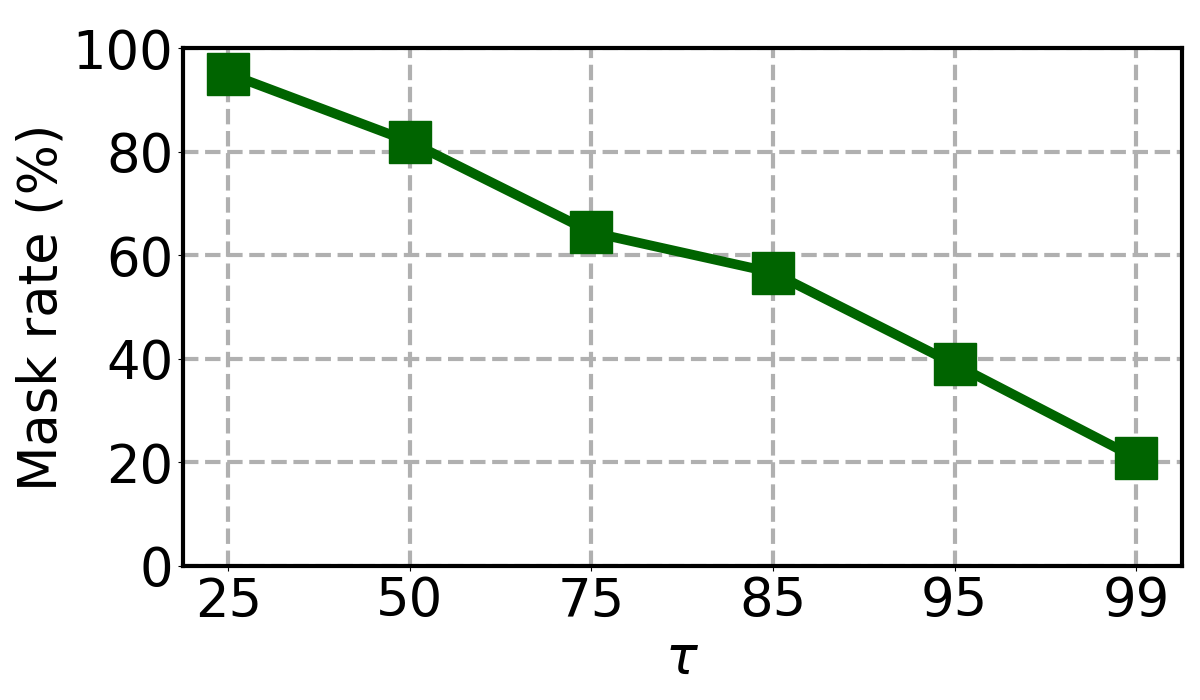}}
\hspace{0cm}
\subfigure[Purity]{
\label{scan_ex} 
\includegraphics[height=2.8cm]{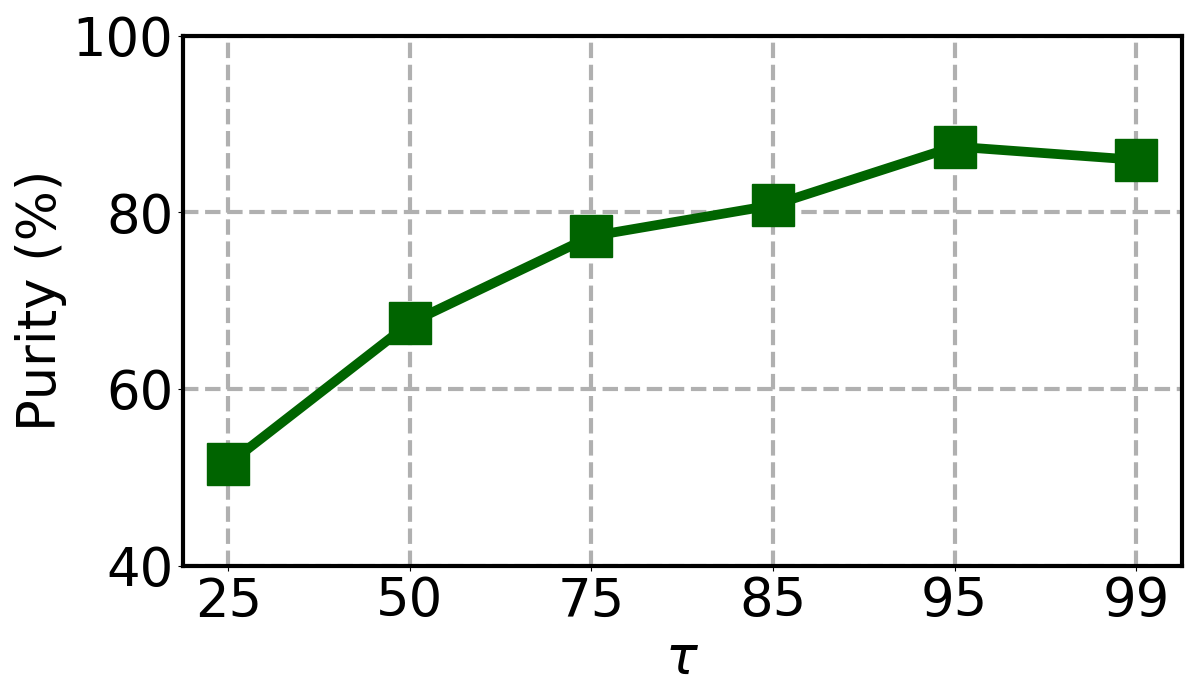}}
\hspace{0cm}
\subfigure[Target accuracy]{
\label{scan_ex} 
\includegraphics[height=2.8cm]{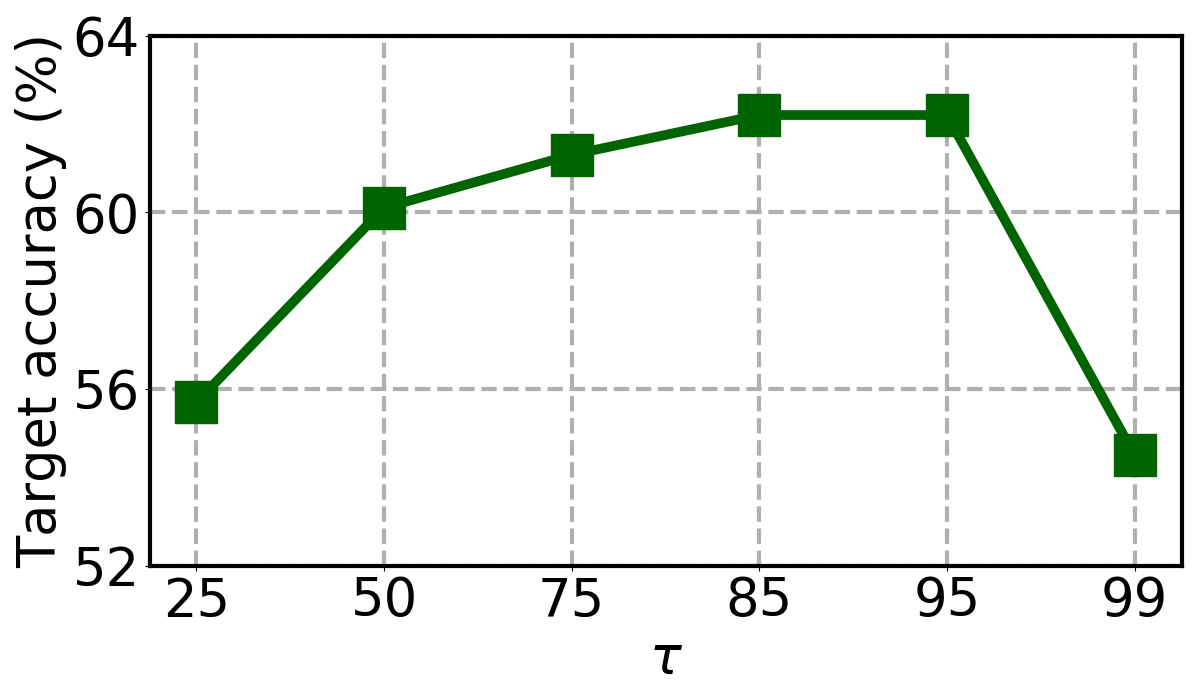}}
\caption{Parameter sensitivity investigation of $\tau$ in terms of (a) mask rate, (b) purity and (c) target accuracy on Oracle-241 dataset.}
\label{sensi_thred} 
\end{figure*}

\textbf{Trade-off paramter $\lambda$.} In our model, the trade-off parameter $\lambda$ is utilized to control the loss of $\mathcal{L}_{d}$. We run the experiments on Oracle-241 and MNIST-USPS-SVHN datasets with the parameter ranges $\lambda\in \left [ 0.1, 0.3,..., 5, 10 \right ] $. As shown in Fig. \ref{sensitivity}, the target accuracy first increases and then decreases as $\lambda$ varies, demonstrating a desirable bell-shaped curve.  A smaller value of $\lambda$ would avoid UDCN optimizing the transition matrices well, which results in the failure of achieving low-density separation between classes. However, an extremely large value of $\lambda$ causes the performance degradation. Especially, the model even fails to converge on the U$\rightarrow $M task when $\lambda \geq 5$. A larger weight on $\mathcal{L}_{d}$ will weaken the effect of source classification loss in Eq. (\ref{source}). Overemphasizing unsupervised learning may lead to the collapse of the network. It can be seen that our UDCN achieves satisfactory results on both Oracle-241 and MNIST-USPS-SVHN datasets with $\lambda = 1$.

\textbf{Confidence threshold $\tau$.} In UDCN, we only assign pseudo labels for the samples whose largest class probabilities $p_i^\alpha$ are larger than $\tau$. To better understand the role of this confidence threshold, we perform parameter sensitivity investigation of $\tau$ in terms of mask rate, purity and target accuracy on Oracle-241 dataset. The mask rate represents the ratio of pseudo-labeled samples to all samples; while the purity denotes the correctness of pseudo-labeling. They can be computed as follows:
\begin{equation}
\text{mask rate} = \frac{1}{n_t}\sum_{i=1}^{n_t}\mathbbm{1}\left ( \max p_i^\alpha> \tau \right ),
\end{equation}
\begin{equation}
\text{purity} = \frac{\sum_{i=1}^{n_t}\mathbbm{1}\left ( \max p_i^\alpha> \tau \right )\mathbbm{1}\left ( \hat{y}_i^t =y_i^t\right )}{\sum_{i=1}^{n_t}\mathbbm{1}\left ( \max p_i^\alpha> \tau \right )},
\end{equation}
where $y_i^t$ is the ground-truth label of $x_i^t$. As shown in Fig. \ref{sensi_thred}, mask rate decreases but purity increases with the increase of $\tau$. When using low threshold values, most target samples are assigned pseudo labels and can be utilized in $\mathcal{L}_{u}$. However, these pseudo labels are always unreliable leading to the error accumulation in training process. Larger $\tau$ forces the network to only assign pseudo labels for the target samples with high confidence, which improves the quality of pseudo labels. Therefore, the target accuracy first increases and then decreases as $\tau$ varies and achieves the best result with $\tau = 0.95$ when both the quality and quantity of pseudo labels are taken into consideration.

\begin{figure*}
\centering
\subfigure[ResNet ]{
\label{visual1} 
\includegraphics[height=3.15cm]{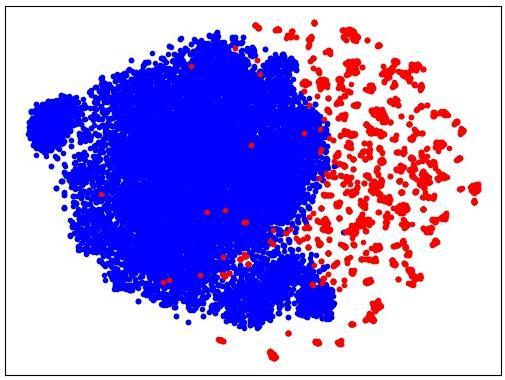}}
\hspace{0cm}
\subfigure[DANN \cite{Ganin2015Unsupervised}]{
\label{visual2} 
\includegraphics[height=3.15cm]{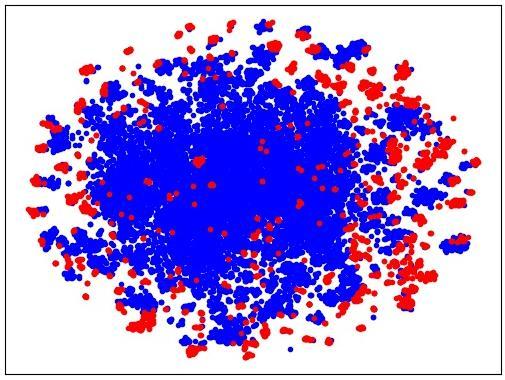}}
\hspace{0cm}
\subfigure[STSN \cite{9757826}]{
\label{visual3} 
\includegraphics[height=3.15cm]{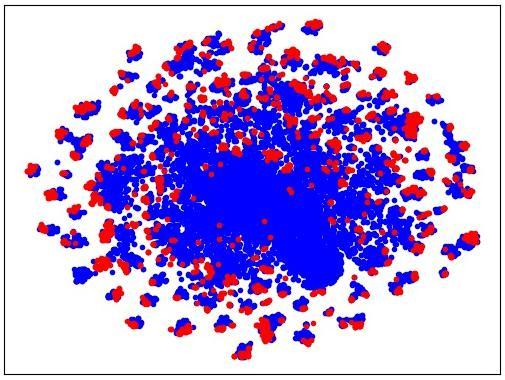}}
\hspace{0cm}
\subfigure[UDCN (ours)]{
\label{visual4} 
\includegraphics[height=3.15cm]{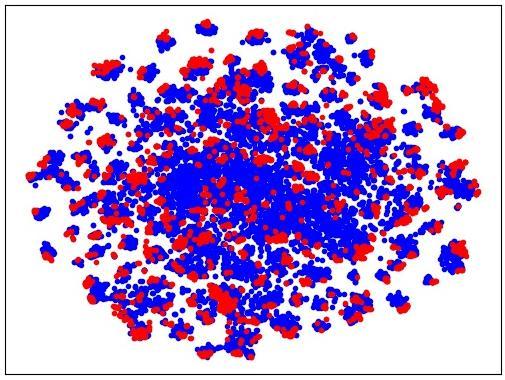}}
\caption{t-SNE embedding visualizations on Oracle-241. Colors denote different domains (red: handprinted data, blue: scanned data).}
\label{visual} 
\end{figure*}

\subsection{Visualization}

\begin{figure}
\centering
\subfigure[Source domain]{
\label{Convergence_a} 
\includegraphics[width=4.25cm]{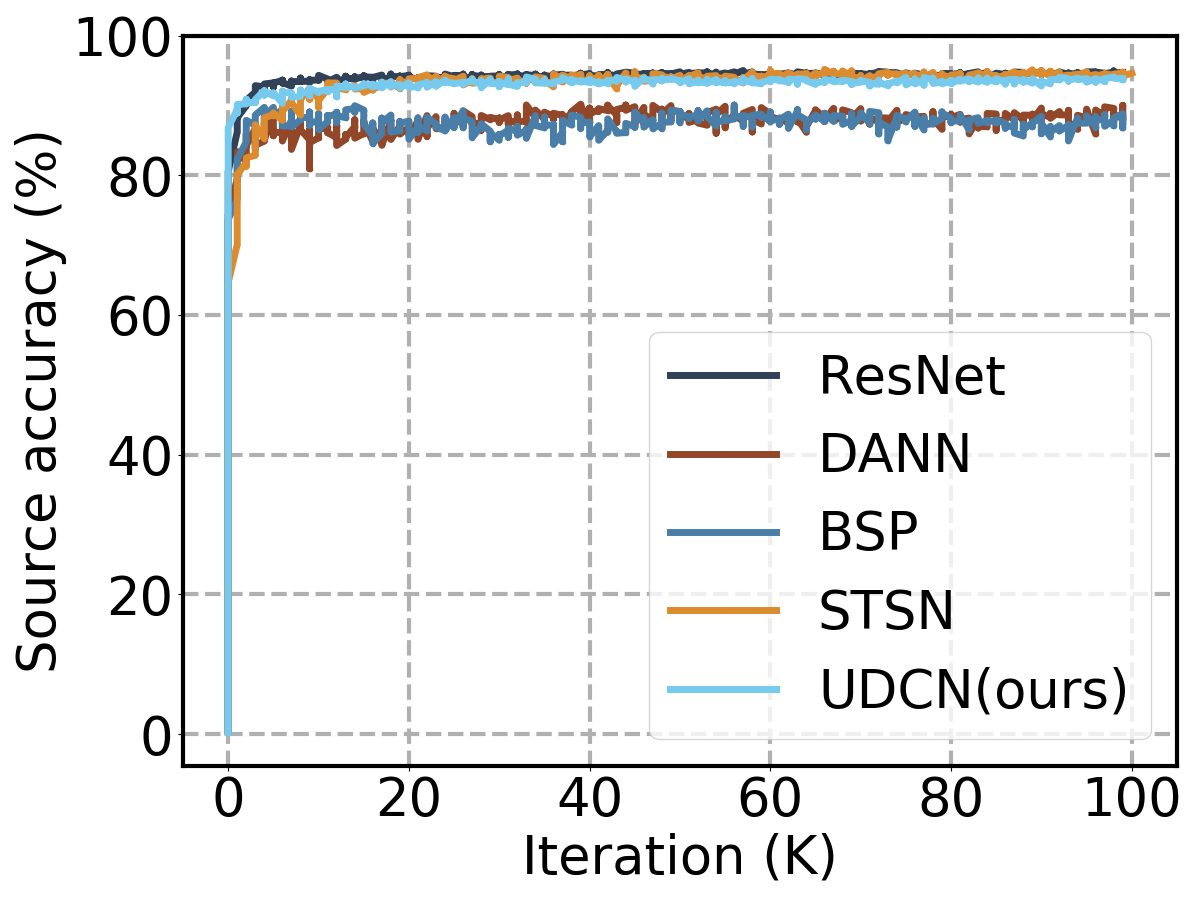}}
\hspace{-0.2cm}
\subfigure[Target domain]{
\label{Convergence_b} 
\includegraphics[width=4.25cm]{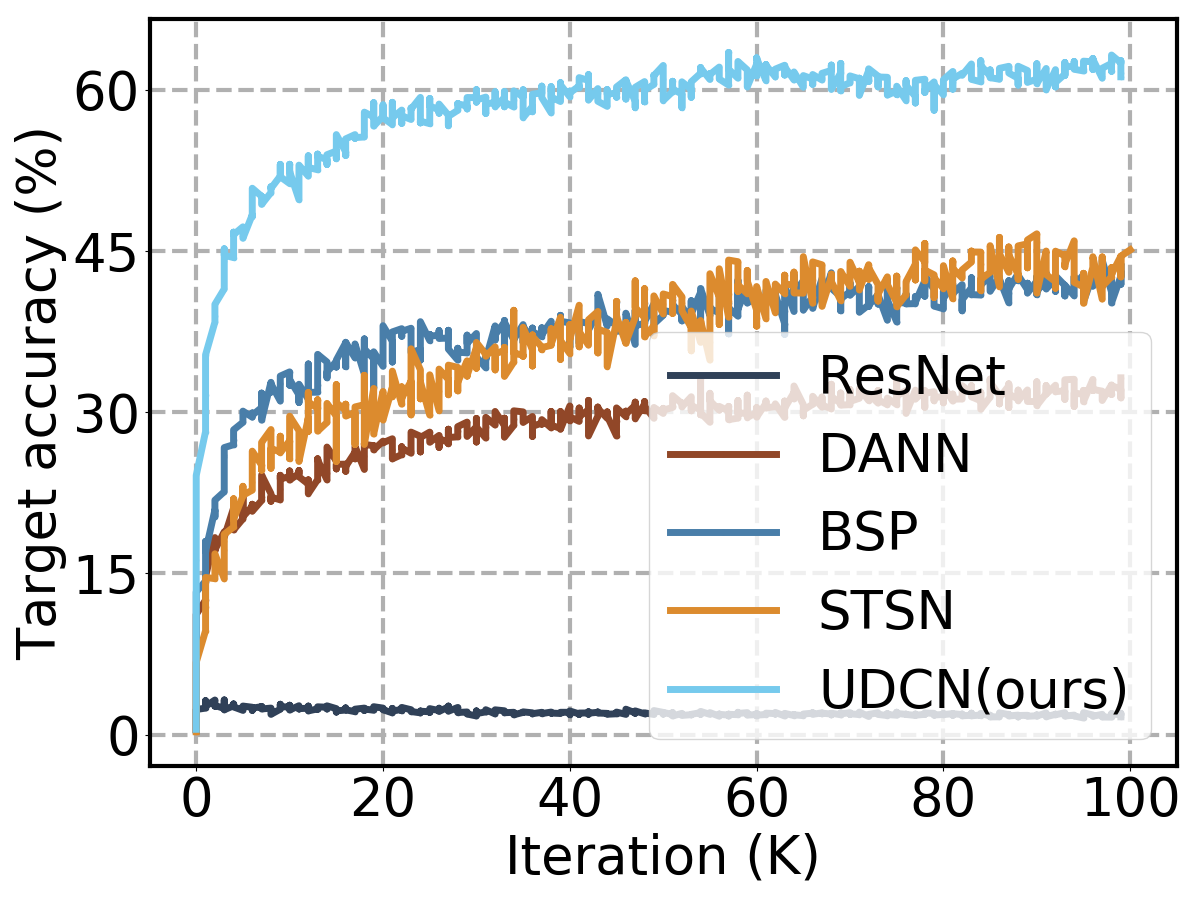}}
\caption{Convergence of ResNet, DANN \cite{Ganin2015Unsupervised}, BSP \cite{chen2019transferability}, STSN \cite{9757826} and our UDCN on the source and target domains of Oracle-241 dataset.}
\label{Convergence} 
\end{figure}

\textbf{Convergence.} We testify the convergence of ResNet, DANN \cite{Ganin2015Unsupervised}, BSP \cite{chen2019transferability}, STSN \cite{9757826} and our UDCN on Oracle-241 dataset. Fig. \ref{Convergence} illustrates the corresponding source and target accuracies during training. From the results, we can find that our UDCN stably converges with the increased number of iteration and shows faster convergence rate than STSN \cite{9757826}. STSN must rely on GAN to generate the transformed scanned data; while the optimization of our proposed method is simple and stable. Moreover, UDCN has much higher target accuracy than STSN, showing the better generalization of our method.

\begin{figure*}
\centering
\subfigure[ResNet ]{
\label{visual_target_1} 
\includegraphics[height=3.15cm]{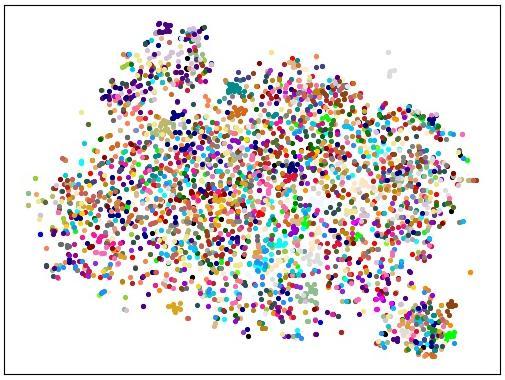}}
\hspace{0cm}
\subfigure[DANN \cite{Ganin2015Unsupervised}]{
\label{visual_target_2} 
\includegraphics[height=3.15cm]{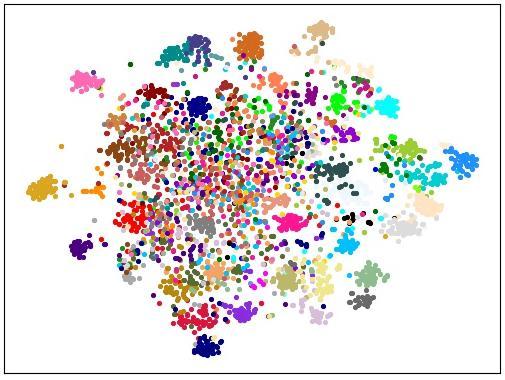}}
\hspace{0cm}
\subfigure[STSN \cite{9757826}]{
\label{visual_target_3} 
\includegraphics[height=3.15cm]{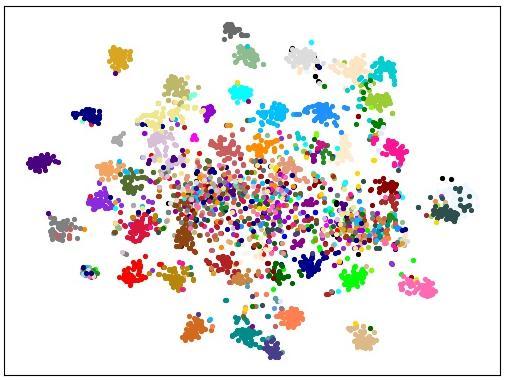}}
\hspace{0cm}
\subfigure[UDCN (ours)]{
\label{visual_target_4} 
\includegraphics[height=3.15cm]{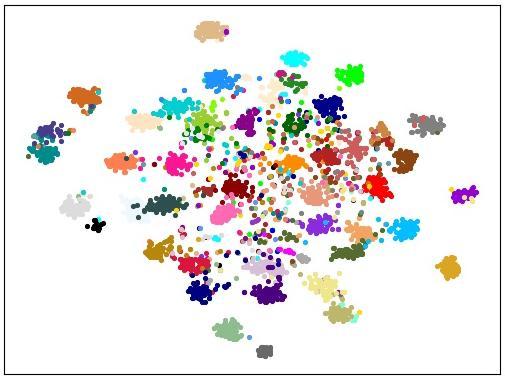}}
\caption{t-SNE embedding visualizations of different target classes on Oracle-241. Colors denote different classes.}
\label{visual_target} 
\end{figure*}

\begin{figure*}
\centering
\subfigure[ResNet ]{
\label{usps2mnist_1} 
\includegraphics[height=3.15cm]{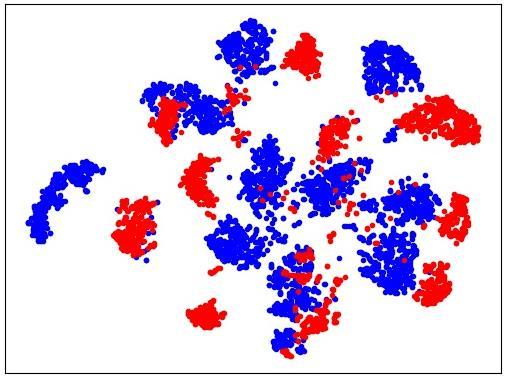}}
\hspace{0cm}
\subfigure[DANN \cite{Ganin2015Unsupervised}]{
\label{usps2mnist_2} 
\includegraphics[height=3.15cm]{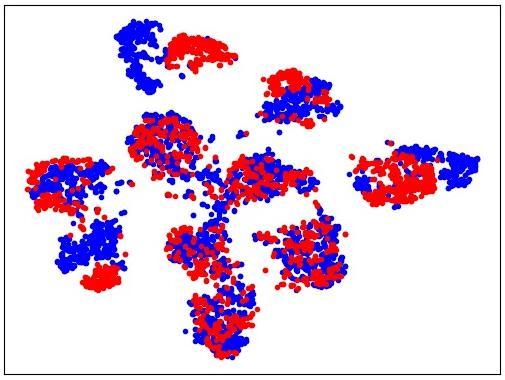}}
\hspace{0cm}
\subfigure[STSN \cite{9757826}]{
\label{usps2mnist_3} 
\includegraphics[height=3.15cm]{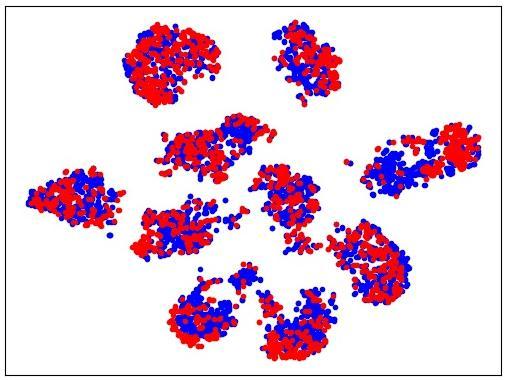}}
\hspace{0cm}
\subfigure[UDCN (ours)]{
\label{usps2mnist_4} 
\includegraphics[height=3.15cm]{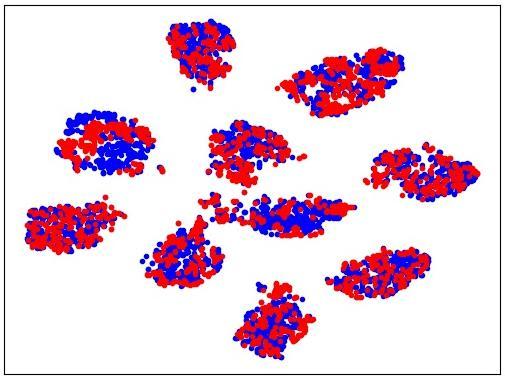}}
\caption{t-SNE embedding visualizations for the U$\rightarrow $M task on digit datasets. Colors denote different domains (red: USPS, blue: MNIST).}
\label{visual_usps2mnist} 
\end{figure*}

\textbf{Feature visualization.} Fig. \ref{visual} shows the t-SNE embeddings of the learned features by ResNet, DANN \cite{Ganin2015Unsupervised}, STSN \cite{9757826} and our UDCN on Oracle-241. Although DANN \cite{Ganin2015Unsupervised} and STSN \cite{9757826} align the two domains and fuse the features, the domain gap still remains in feature space. Our UDCN can totally mix up the source and target features and achieve satisfactory alignment. Moreover, to evaluate the effectiveness of UDCN on learning discriminative target features, we also randomly select some target samples from 60 different classes of Oracle-241 and show their embeddings using t-SNE in Fig. \ref{visual_target}. Although DANN \cite{Ganin2015Unsupervised} and STSN \cite{9757826} improve the model performance on scanned data, most samples with different class labels are still mixed together. In our UDCN, samples in the same class are pulled closer and samples with different classes are dispersed. It demonstrates that UDCN successfully passes the class information to the network and reduces class confusion leading to more discriminative features. Similar observation can be obtained on digit dataset shown in Fig. \ref{visual_usps2mnist}.

\begin{figure}
\centering
\includegraphics[width=8.5cm]{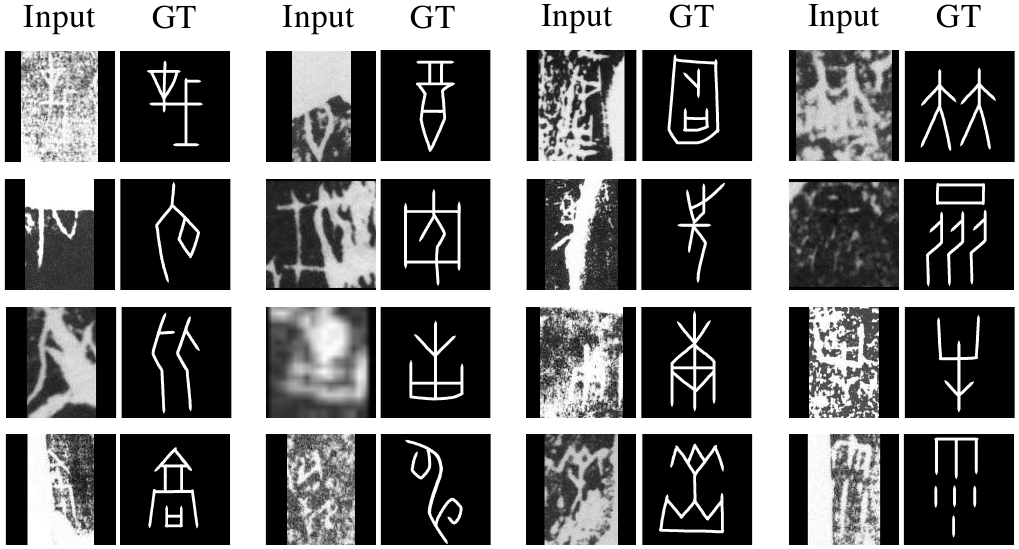}
\caption{Some examples which are misclassified by STSN \cite{9757826} but classified correctly by our UDCN. The ground-truths are showed by the corresponding handprinted images belonging to the same classes.}
\label{error} 
\end{figure}

\textbf{Effectiveness on hard samples.} We compare the performance between STSN \cite{9757826} and UDCN on some severely abrasive characters. The results are shown in Fig. \ref{error}. We have two main observations. (1) Although the recently proposed STSN method obtains the second best result in terms of target accuracy shown in Table \ref{oracle-241}, it still fails to recognize some severely abrasive characters. (2) Our UDCN successfully improves the model robustness to abrasion and noise, and correctly classifies these hard samples which are even confused for humans.

\section{Conclusion}

In this work, an unsupervised discriminative consistency network (UDCN) was proposed for UDA of oracle character recognition. To handle the serious abrasion and noise in real-world oracle characters, we proposed to enhance the pseudo-label based consistency across different augmented views and thus improved the model robustness. Moreover, when considering the problem caused by various writing styles and inter-class similarity, we designed an unsupervised transition loss based on the transition matrix to reduce the correlation between different classes and learn more discriminative features. Comprehensive comparison experiments on Oracle-241 and MNIST-USPS-SVHN datasets strongly demonstrated the state-of-the-art performance of our UDCN.

However, several limitations, including future works, need to be addressed. First, the intrinsic properties of oracle characters, e.g., radicals, are not taken advantage of in UDCN, which remains to be explored. Second, UDCN mainly focuses on close-set UDA of oracle character recognition. Our further research will delve into extending our method to more complicated distribution shift problems, e.g., partial UDA and open-set UDA.

\section{Acknowledgements}	

This work was supported by China Postdoctoral Science Foundation under Grant 2022M720517 and National Natural Science Foundation of China under Grant 62236003 and 62306043.

{
\bibliographystyle{IEEEtran}
\bibliography{egbib}
}

\end{document}